\RequirePackage{fix-cm}
\documentclass[smallextended]{svjour3}       
\smartqed  
\usepackage[english]{babel}             
\usepackage{mathptmx}       		     
\usepackage{dsfont}					
\usepackage{amssymb}				
\usepackage{amsmath}				
\usepackage{graphicx}				
\usepackage{sidecap}				
\usepackage[inline]{enumitem}           
\usepackage[hidelinks]{hyperref}        
\usepackage{pgfplots}				
\usepackage{hhline}					
\usepackage{multirow}				
\usepackage{diagbox}				
\usepackage{rotating}				
\usepackage{caption}				
\usepackage{bold-extra}                 
\usepackage{subfig}                     
\sidecaptionvpos{figure}{c}

\newcommand{\dom}[1]{\ensuremath{\mathds{#1}}}

\newcommand{\sta}{\ensuremath{s}}
\newcommand{\st}[1]{\ensuremath{\sta_{#1}}}%

\newcommand{\tim}{\ensuremath{t}}
\newcommand{\tm}[1]{\ensuremath{\tim_{#1}}}%
\newcommand{\tmWR}[1]{\ensuremath{\text{?}\tm{\onto{#1}}}}
 
\newcommand{\STA}[1]{\ensuremath{#1}}
\newcommand{\STAs}[2]{\ensuremath{\STA{#1}^{#2}}}

\newcommand{\g}{\ensuremath{\textbf{\onto{g}}~}}

\newcommand{\swrlG}[1]{\ensuremath{\text{swrl}_{>}(#1)}}
\newcommand{\swrlL}[1]{\ensuremath{\text{swrl}_{<}(#1)}}
\newcommand{\swrlA}[1]{\ensuremath{\text{swrl}_{+}(#1)}}

\newcommand{\onto}[1]{\ensuremath{\texttt{#1}}}

\newcommand{\hasSt}[1]{\ensuremath{\onto{hasState(}#1\onto{)}}}
\newcommand{\hasTm}[1]{\ensuremath{\onto{hasTime(}#1\onto{)}}}

\newcommand{\TImporter}[1]{\ensuremath{\onto{w\textsubscript{#1}}}}
\newcommand{\ModelOnto}[1]{\ensuremath{\onto{o\textsubscript{#1}}}}
\newcommand{\TDetecter}[1]{\ensuremath{\overline{\onto{w}}\textsubscript{\onto{#1}}}}
\newcommand{\PlaceOnto}{\ModelOnto{0}}
\newcommand{\DImporter}{\TImporter{0}}

\usepackage{xcolor}
\newcommand{\coloredBox}[1]{{\color{#1} \rule{0.045\linewidth}{2mm} }}
\definecolor{Leg1}{RGB}{0,80,250}
\definecolor{Leg2}{RGB}{255,0,20}
\definecolor{Leg3}{RGB}{0,250,70}
\definecolor{Leg4}{RGB}{190,100,60}
\definecolor{Leg5}{RGB}{30,130,130}
\definecolor{Leg6}{RGB}{255,70,250}
\definecolor{Leg7}{RGB}{0,255,255}
\definecolor{Leg8}{RGB}{0,0,0}
\definecolor{Res1}{RGB}{255,0,0}
\definecolor{Res2}{RGB}{10,250,70}
\definecolor{Res3}{RGB}{0,60,250}
\definecolor{Res4}{RGB}{0,0,0}
\definecolor{topGreen}{HTML}{008000}
\definecolor{topRed}{HTML}{A02C2C}

\usepackage{tikz}
\usepackage{rotating}

\begin{document}\sloppy

\title{Towards a new paradigm for assistive technology at home: research challenges, design issues and performance assessment  
}

\titlerunning{Towards a new paradigm for assistive technology at home} 

\author{Luca Buoncompagni \and Barbara Bruno \and Antonella Giuni \and Fulvio Mastrogiovanni \and Renato Zaccaria
}

\authorrunning{Luca Buoncompagni et al.} 

\institute{
L. Buocompagni \at
Department of Informatics, Bioengineering, Robotics and Systems Engineering\\
University of Genoa, Italy\\
\email{luca.buoncompagni@edu.unige.it}
\and 
B. Bruno \at
Department of Informatics, Bioengineering, Robotics and Systems Engineering\\
University of Genoa, Italy\\
Teseo srl\\
\email{barbara.bruno@unige.it}
\and 
A. Giuni \at
Teseo srl\\
\email{antonella.giuni@teseotech.com}
\and
F. Mastrogiovanni \at
Department of Informatics, Bioengineering, Robotics and Systems Engineering\\
University of Genoa, Italy\\
Teseo srl\\
\email{fulvio.mastrogiovanni@unige.it}
\and
R. Zaccaria \at
Department of Informatics, Bioengineering, Robotics and Systems Engineering\\
University of Genoa, Italy\\
\email{renato.zaccaria@unige.it}
}

\date{Received: date / Accepted: date}

\maketitle

\begin{abstract}
Providing elderly and people with special needs, including those suffering from physical disabilities and chronic diseases, with the possibility of retaining their independence \textit{at best} is one of the most important challenges our society is expected to face. 
Assistance models based on the home care paradigm are being adopted rapidly in almost all industrialized and emerging countries.
Such paradigms hypothesize that it is necessary to ensure that the so-called Activities of Daily Living are correctly and regularly performed by the assisted person to increase the perception of an improved quality of life.
This chapter describes the computational inference engine at the core of Arianna, a system able to \textit{understand} whether an assisted person performs a given set of ADL and to \textit{motivate} him/her in performing them through a speech-mediated motivational dialogue, using a set of \textit{nearables} to be installed in an apartment, plus a wearable to be worn or fit in garments.

\end{abstract}

\section{Introduction}
\label{section:introduction}

During the past few years, different approaches to allow elderly and people with special needs to retain their independence as long as possible, while living at home, have been pursued both in academic research and as part of product-oriented design and development efforts.  
As a reference, the Ambient Assisted Living (AAL) market is valued today approximately \$1 billion, with a 55.6\% CAGR over the 2017-2021 period.

Assistance models based on the home care paradigm are being adopted rapidly in almost all industrialized and emerging countries.
It addresses two intertwined needs:
(i) supporting elderly and people with special needs, both in a post-hospitalization phase and when it is necessary to have a personalized mid- or long-term care service;
(ii) helping people who do not have an easy access to hospital-based services, e.g., because they live in the countryside.
In both cases, the home care paradigm assumes that it is necessary to ensure that the so-called Activities of Daily Living (ADL) are correctly and regularly performed by the assisted person to increase the perception of an improved quality of life.
ADL are daily activities related to motion, rest, nutrition, and personal hygiene, which are a qualitative indicator of a person's wellbeing, determine their quality of life and level of independence.

This chapter describes the computational inference engine at the core of Arianna, a system able to \textit{understand} whether an assisted person performs a given set of ADL and to \textit{motivate} him/her in performing them through a speech-mediated motivational dialogue, using a set of \textit{nearables} to be installed in an apartment, plus a wearable to be worn or fit in garments.
Arianna originates from joint work carried out by University of Genoa and Teseo srl.
The ideas underlying Arianna are based on new approaches to the management of chronic diseases such as cognitive decline \cite{Bredesen2014,Bredesen2016}, i.e., adopting personalized and multi-therapeutic approaches.
ADL are particularly relevant to such treatments.
These studies show that adopting a proper lifestyle is an essential step in the management and in some cases the regression of disabling chronic diseases.

A number of functional requirements are expected from such a system:
\begin{itemize}
\item localizing people in their home or apartment;
\item identifying their significant gestures and correlating them to position and time of day;
\item determining people activities related to ADL;
\item interacting with the assisted person through dialogues;
\item reminding people, by means of voice interaction, to perform typical ADL, if not detected or performed too rarely, acting as a personal assistant;
\item checking their posture, to allow for a quick intervention in case of falls or fainting;
\item learning their habits and identifying anomalous situations;
\item automatically notifying anomalous situations to relatives, friends, or medical staff. 
\end{itemize}

In particular, the contribution of this chapter is a discussion about the design and the implementation choices related to a semantic model (and the associated inference engine) able to perform multiple human activity recognition and classification, which has been designed to meet soft real-time requirements in real-world use cases.
In particular, we introduce the general definition of an fluent model for human activity recognition and we ground it in a state-of-the-art, rule-based, semantic language, which is implemented in an ontology-based framework.

The chapter is organized as follows.
Section \ref{sec:related_work} discusses relevant background.
Section \ref{sec:arianna_computational_framework} introduces the main ideas underlying the Arianna's computational framework.
Reasoning models are introduced in Section
\ref{sec:grounding}, whereas Section \ref{sec:MON} describes how the system is structured in a network of interacting ontologies. 
Experimental validation is discussed in Section \ref{sec:experiments}.
Conclusions follow.

\section{Related Work}
\label{sec:related_work}

\subsection{Background}
\label{sec:Background}

From the very beginning, technological solutions for the automatic monitoring of metrics related to the well-being of a person have primarily focused on Activities of Daily Living \cite{Liu16}.

ADL are daily-life activities which, when successfully performed, guarantee and imply a certain level of independence. 
Starting from the late $1950$s gerontologists, i.e., scientists studying the social, psychological and biological aspects of aging, have studied people's performance in their execution to analyze the correlation between human actions and motor and cognitive abilities.
The definition of ADL comes from the \textit{Index of Activities of Daily Living} \cite{Katz59}, which introduces a classification of the functional status of elderly people on the basis of their ability in carrying out $6$ activities, i.e., the ADL.
Such ADL include \textit{bathing}, \textit{dressing}, \textit{toileting}, \textit{transferring}, \textit{continence} and \textit{feeding}.

Subsequent research has expanded the set of daily-life activities relevant for the assessment of a person's wellbeing.
The \textit{Scale of Instrumental ADL} \cite{Lawton69} considers $9$ daily activities, which require a certain level of planning capabilities and social skills, such as the usage and interaction with devices of common use.
Proposed Instrumental ADL (IADL) include \textit{placing a telephone call}, \textit{shopping}, \textit{preparing food}, \textit{housekeeping}, \textit{doing the laundry}, \textit{moving outdoor with public transports}, \textit{moving indoor on foot}, \textit{taking medications} and \textit{handling finances}.
Nowadays, the Index of ADL and the Scale of IADL are the \textit{de facto} standard indexes for the assessment of a person's functional status \cite{Bruno14} and the name ADL is often used to collectively denote all considered daily activities.

The next Section provides an overview of commonly adopted approaches designed to monitor ADL and related metrics, such as \textit{motility}, \textit{falls}, and \textit{sleep disorders}.
Information about such metrics can be used to complement the information obtained by the analysis of the ADL, and it can be especially useful in the context of multidimensional geriatric assessments \cite{Rubenstein90,Pilotto08}, which rely on many, often interdisciplinary diagnostic tools to define an elderly individual's medical, psychosocial, and functional capabilities and problems to arrive at an overall plan for therapy and long-term follow-up.

The survey in the next Section has a technological, development-oriented perspective.
To better contextualize the presented work and the state-of-the-art, we argue that a tight connection with the application, i.e., with the healthcare requirements driven by the aforementioned goals, is of the utmost importance.
To this aim, it is interesting to review the main findings of a survey of smart homes and home health-monitoring solutions done from a medical, application-oriented perspective \cite{Liu16}.
Of the $6$ key points therein raised, $3$ concern the evaluation of smart homes and health-monitoring technologies, to collect clear evidence about the \textit{effectiveness}, \textit{sustainability} and \textit{usability} of such solutions in real use conditions.
The other $3$ points concern the design of such solutions, with a particular focus on the \textit{cost} for the final user, which has an obvious strong impact of the \textit{marketability} and \textit{sustainability} of the solution, and especially for developing countries and disadvantaged population sections.

The proposed solution takes these insights into account to propose a number of key functional capabilities, which must be considered as a sort of baseline when monitoring ADL:
\begin{itemize}
\item monitoring people locations in the environment at the \textit{topological} level, e.g., \textit{in the kitchen}, \textit{in the bedroom}, \textit{in the bathroom};
\item monitoring postures and transitions between postures, e.g., \textit{standing}, \textit{seated}, \textit{sitting down}, and \textit{falls};
\item monitoring a selected number of important ADL based on gesture detection, e.g., \textit{walking}, \textit{drinking}, \textit{using fork and knife}, \textit{teeth brushing};
\item speech-based dialogues between Arianna and the assisted person to actively obtain information about his/her state, e.g., \textit{determining if an unusual motion pattern corresponds to a sudden illness};
\item motion analysis and detection of special sequences of activities, e.g., \textit{sitting for a long time}, \textit{staying still for a long time}, \textit{cooking}, \textit{going often to the bathroom}.
\end{itemize}


\subsection{Human Activity Recognition}
\label{sec:survey}

A large corpus of literature deals with the automatic recognition of ADL.
Beside differences in the adopted sensing equipment and, consequently, in the middleware for the management of the sensory data and in the techniques for their analysis, most solutions follow the same working principle of ``distributed sensing, centralized reasoning''.
In this paradigm, (possibly) heterogeneous raw sensory data are collected and, with none or minimum pre-processing, made available to the reasoning system responsible for analyzing them to extract high-level information related to the execution of ADL.
As an example, the richest among such solutions deploy large numbers of inexpensive, binary sensors (such as presence sensors to detect whether there is a person in an area, open/closed sensors for doors and windows, on/off sensors for light switches, etc.) and rely on \textit{ontologies} for the extraction of high-level ADL-related information from sensory data \cite{Cook13}\cite{Scalmato13}, with no intermediate processing.
The daily life activities that can be monitored with such solutions include \textit{food preparation}, \textit{feeding}, \textit{indoor transportation}, \textit{toileting} and \textit{using communication devices} such as telephones \cite{Cook13}\cite{Scalmato13}, possibly complemented with additional information, for example concerning leisure activities (\textit{resting on the sofa}, \textit{watching TV}) or sleep quality \cite{Scalmato13}.

The reliability of the recognition is, obviously, a key requirement for all such systems, and among the enabling factors for a reliable solutions there is the ability of discriminating between (and appropriately notifying) situations where all sensory data are consistent with the recognized activity and others where the recognition is affected by missing or contradicting information.
In this respect, one solution for the monitoring of food preparation and toileting activities is worth a special attention \cite{Hong09}.
As in the above solutions, the proposed system relies on a network of distributed binary sensors (presence sensors complemented with pres
sure mats for inferring the person's location and contact switches for detecting whenever a door/shutter is opened).
The reasoning module extends a hierarchy of coarse-to-fine ontologies with methods implementing the Dempster - Shafer (DS) theory of evidence, which allows for providing rich information about the confidence of the recognition.

Beside such broad solutions, a number of systems have been developed specifically aimed at maximizing the recognition accuracy of one type of ADL only.
Solutions following this paradigm usually rely on a single or very few homogeneous sensors, which sense quantities inherently related to a narrow range of activities.
As an example, by solely analyzing the water fixtures in the house it is possible to accurately detect when a person is bathing, toileting, or doing the laundry \cite{Thomaz12}\cite{Hu13}.

A number of solutions aim at the analysis of the walking pattern of people inside their homes, which, beside being related to the ADL \textit{indoor transportation} and a relevant metric \textit{per se}, allows for enhancing the recognition of other daily-life activities as well (e.g., knowledge about the current position of the person within the home can help with the recognition of all those activities occurring at specific locations, e.g., bathing or cooking).
The detection and tracking of a person's location is usually done under one of two approaches, i.e., either with:
i) a network of distributed presence sensors, signaling whenever there is movement in the sensed area \cite{Boers09};
or ii) a localization system which relies on radio signals and triangulation/trilateration algorithms to determine the position of a mobile device, usually called \textit{tag}, with respect to fixed devices distributed in the environment in known positions, called \textit{beacons} or \textit{anchors} \cite{Nepa10}\cite{Kim13}.
The second approach allows for estimating the location of the monitored person with a higher accuracy than the first one and, unlike the first one, is robust to multiple people living in the same home or apartment. 
Nonetheless, by requiring the monitored person to physically carry around the tag device, this approach negatively impacts the person's motion freedom and appearance.

Lastly, a number of solutions focus on developing a low-cost, non-intrusive technology for providing an external observer with the tools for extracting basic, but accurate information about the house inhabitant's daily activities, especially targeting the identification of anomalous behaviors or events. 
As an example, one of the simplest anomaly detection systems in the literature monitors the TV operational status (on/off), since its pattern expresses the consistency of long-established everyday rituals and is particularly apt for the detection of deviations from the typical behavior \cite{Nakajima08}.
Another solution expands the analysis to all electrical devices in the house via a plug-in sensor able to detect the electrical noise generated on residential lines by the operational modes of electrical devices \cite{Gupta10}.
The system autonomously identifies the device generating the noise by matching the detected interference with the ``fingerprint'' interference of all the devices of interest, recorded in the setup phase.
In accordance with the same principle, other authors analyze the water pressure within the house water infrastructure to identify individual water fixtures according to the unique pressure waves that propagate to the sensor when the valves are opened or closed \cite{Froehlich09}, while a last example uses air pressure sensors attached to the Heating, Ventilation, Air Conditioning system of a house to detect doorway transitions \cite{Patel08}.

A few considerations are of relevance.
First of all, to the best of our knowledge, no state-of-the-art solution allows for the recognition of all the ADL and metrics of interest discussed in Section \ref{sec:Background}.
Secondly, there is a many-to-many mapping between the ADL and the recognition solutions (both from a hardware and software perspective), i.e., there exist complementary approaches for the monitoring of the same ADL (e.g., for the detection of toileting activities) and, at the same time, the same monitoring solution can be used to provide information, possibly at different levels of detail and accuracy, about a number of ADL (e.g., presence sensors).

The second consideration has two important implications:
on the one hand, it makes it more difficult to identify the ``best'' monitoring solution for given conditions,
while on the other hand allows for \textit{adaptable} solutions, which can suit different monitoring requirements, personal constraints and house settings.
Moreover, the number of state-of-the-art solutions for monitoring ADL-related variables, as well as the rate at which new hardware is introduced and new approaches are developed, suggests that even better performance could be achieved by combining different solutions in a \textit{modular} framework, under the principle of ``distributed sensing, distributed reasoning''.
Solutions under this paradigm should:
\begin{itemize}
\item maintain a high-level uniform interface towards the caregivers, regardless of the specific adopted monitoring solutions;
\item enforce the independence of the specific monitoring solutions, to allow for the integration of any hardware and software approach, with minimum effort;
\item be \textit{scalable} with respect to the number of hardware and software modules, to allow for real adaptability (i.e., any combination of hardware and software monitoring solutions).
\end{itemize}


\section{The Arianna Computational Framework}
\label{sec:arianna_computational_framework}

On the basis of the considerations put forth in the previous Section, this work presents a computational framework to describe and evaluate models for human activity recognition, which has been designed to be flexible, modular and scalable in terms of the number of activities to be monitored, the associated  techniques and their \textit{meaning}, as well as to the different environments they can occur within.
In particular, we discuss the activity modeling formalism used in Arianna, and the results of a number of tests carried out in simulation, which focus on its computational complexity, both from the technological (i.e., hardware and software) as well as the caregiver interface perspectives.

These features are enforced via the use of a Description Logic formalism \cite{DL}, i.e., a standard, state-of-the-art framework for what concerns natural data representation, contextualization and semantic reasoning.
Any interface towards caregivers results to be simplified with respect to what would be possible with other formalisms, whereas the expressibility of available reasoners provides the system with an extended flexibility as far as inferences and deductions are concerned.
Nevertheless, semantic reasoning is a resource consuming task.
Therefore, we focused system's validation on analyzing and optimizing the overall system's performance.
This lead us to a system's design where a \textit{network} of ontologies is tasked with a distributed reasoning process, each one dealing with different levels of detail in knowledge representation.

Formalization of activity modeling has been carefully designed with two purposes in mind:
i) enforcing flexibility and modularity in the framework as far as reasoning is concerned, and
ii) achieving a scalable system by proposing common design patterns for developing human activity recognition modules.

A fundamental step to achieve such a distributed reasoning system is the definition of an activity model, its components and the procedures to evaluate it.
In particular, Arianna describes each data sample through a fluent \emph{statement} defined as a combination of a Boolean state (e.g., \textit{top} $\top$ and \textit{bottom} $\bot$) and a generation time instant (henceforth indicated by a capital letter).
In order to compose such statements we define a set of logical operations (e.g., the \textit{is-a} relationship $\sqsubseteq$, the \textit{conjunction} $\sqcap$), concerning their state, as well as algebraic operations (e.g., $+$, $\geqslant$) addressing their relations over time.
Through such a formalism we define a \emph{model} as a bijective function generating a new statement with a true state if and only if certain logical and/or algebraic relations hold. 

As an example of a fluent model, let us consider the problem of recognizing an activity consisting in picking two objects from a cabinet, using them for a while and finally placing them back in the cabinet.
In particular, let us consider a smart environment where those objects are provided with sensors able to generate statements \STA{I} and \STA{O} with state $\top$ if the object is in the cabinet, $\bot$ otherwise.
Moreover, let us assume a sensor attached to the cabinet door (\STA{D}) with state $\top$ if it is open and $\bot$ otherwise.
Such a model assumes two statements, namely object taken (\STA{T}) and released (\STA{R}), such that the activity is considered to be accomplished when those statements hold true after a certain, minimum delay $\delta$ between each other (i.e., the objects have been used for a while).
It is noteworthy that the recognition of the activity is represented as a new belief (i.e., a statement \STA{A}), which is $\top$ if it has been performed and it specifies also the time when activity execution is inferred.
Let us analyze how the models to generate those three statements can be formalized. 
The objects are considered to be \emph{taken} when the cabinet door is open and then the items are absent:
\begin{equation}
(\STAs{D}{\top} < \STAs{I}{\bot}) \sqcap (\STAs{D}{\top} < \STAs{O}{\bot}) \Rightarrow \STAs{T}{\top}
\label{eq:m-exT}
\end{equation}
In a similar way, we can define the fact that the items are \emph{released} if the cabinet door is closed and, after that, the fact that the items are present:
\begin{equation}
(\STAs{D}{\bot} > \STAs{I}{\top}) \sqcap (\STAs{D}{\bot} > \STAs{O}{\top}) \Rightarrow \STAs{R}{\top}
\label{eq:m-exR}
\end{equation}
Finally, the model considers that the activity is completed and generates an \STA{A} belief if the objects have been taken and released after $\delta$ time units,
\begin{equation}
(\STAs{T}{\top} + \delta) < \STAs{R}{\top} \Rightarrow \STAs{A}{\top}\label{eq:m-exF}
\end{equation}

It is noteworthy that more complex models can be built as chains of sub-models, since their outcome are statements, which can be embedded in more complex statements.
This makes the Arianna reasoning engine scalable and suitable to validate different ways to assert the same abstract concepts (e.g., the case of \textit{object taken}). 
The only constraint we assign to a model formula is the use of logical operators to aggregate states and algebraic relations for their temporal classification, but we do not limit the complexity of their functions.

\section{Model Grounding and Reasoning}
\label{sec:grounding}

\subsection{Rationale}
\label{sec:rationale}

Let us consider a system that dynamically evaluates several models within a semantic data representation able to reason and create inferences based on available statements.
In this case, observations (e.g., \textit{the object has been taken}) are considered as beliefs shared with other models, which truth values may change based on such beliefs and other contextualized statements (i.e., reasoning outcomes). 
For instance, the fact that \emph{some kitchen items have been used} is a suitable input statement for a model detecting an eating activity, e.g., it may be used to create an \emph{had breakfast} belief if it were detected in the morning.
The same statement may be useful for other models, for instance to generate an alert belief if the same items were used during nighttime. 

The Ontology Web Language \cite{OWL} well describes such characteristics by logically define appropriate semantics, so that their availability make the reasoner infer new statements to be used as input beliefs for further models.
Those models may generate other contextualized statements as well, with the purposes of updating the overall data structure over time. 

In Arianna, we ground the above statement algebra formalization through a DL-based framework, with the support of the Semantic Web Rule Language (SWRL) \cite{swrl}.
DL-based languages allow for the specification of the \onto{Statement} domain by restricting its instances to have exactly one property specifying its state and another specifying the generation time:
\begin{equation}
\onto{X} \sqsubseteq_{=1} \hasSt{ \st{\onto{X}}} \sqcap_{=1} \hasTm{ \tm{\onto{X}}} \doteq \onto{Statement}
\label{eq:sem-st}
\end{equation}
where -- it is interesting to note -- only a subset of properties of \onto{X} are specified, and other properties of different types, possibly used by the reasoner for further semantic contextualization, may be present. 
Furthermore, it is important to highlight that due to the reasoning processes, performed using the OWL API \cite{OWL_api} and Pellet \cite{pellet} specifications, it is possible to rely on \emph{consistency checking} to describe the context as a semantic hierarchy to be applied to statements over time.
In particular, it is possible to define semantic classes (indicated using capitalized names) and properties in their definitions (indicated with \emph{be/have} characteristics), in order to allow the reasoner to classify individuals based on facts that \emph{have to hold true} (in consistency terms) in the description of the environment.
For instance, a property \onto{hasConfidence} could be used for classifying \onto{X} as an instance of the classes \onto{Unrealistic}, \onto{Probable} or \onto{Accurate}.

The SWRL formalism allows for evaluating conjunctions of facts in order to infer new properties in the ontology, denoted by the symbol $\rightharpoonup$.
Due to this capability, such a language is suitable to ground model definitions.
In particular, rule specification is based on variables, denoted by the `?' symbol, and on built-in logic functions to check semantic types (i.e., instances, properties or classes) as well as simple algebraic expressions.
In the reasoning process, such rules are evaluated through instance checking by replacing all the variables with the specified instances, without a specific priority. 

As a proof-of-concept example of how SWRL rules can be used in this case, let us consider again the model described in Equations \ref{eq:m-exT}, \ref{eq:m-exR} and \ref{eq:m-exF}.
First, in order to simplify the notation, let us define a term substitution $\g(X)$ in charge of retrieving the statement information \st{X} and \tm{X} from the structure, by making them available for further SWRL rules:
\begin{equation}
\g(\STAs{X}{\bar{s}}) ~:~ \hasSt{\onto{X}, \bar{s}} \sqcap \hasTm{\onto{X}, \tmWR{X}}
\end{equation}
Then, it is possible to describe the model to assess whether an object has been taken (\autoref{eq:m-exT}) through two SWRL rules:
\begin{align}
\label{eq:sw-exT}
&\g(\STAs{D}{\top}) \sqcap \g(\STAs{O}{\bot}) \sqcap \g(\STAs{I}{\bot}) \sqcap \swrlL{\tmWR{D}, \tmWR{I}} \sqcap \swrlL{\tmWR{D}, \tmWR{O}}\notag\\
&\sqcap\begin{cases}
\swrlG{\tmWR{I}, \tmWR{O}} & \rightharpoonup \hasSt{\onto{T}, \top} \sqcap \hasTm{\onto{T}, \tmWR{I}}\\
otherwise & \rightharpoonup \hasSt{\onto{T}, \top} \sqcap \hasTm{\onto{T}, \tmWR{O}}
\end{cases}
\end{align}
where $\swrlL{a,b}$ is $\top$ \textit{iff} $a < b$, and analogously for other algebraic operations (e.g., \swrlA{a,b,c} is $\top$ \textit{iff} $c = a + b$).
Furthermore, it is possible to implement the model introduced in \autoref{eq:m-exR}, in order to assert that the objects have been released in their proper location.
This can be done with a single SWRL rule as:
\begin{align}\label{eq:sw-exR}
&\g(\STAs{D}{\bot}) \sqcap \g(\STAs{O}{\top}) \sqcap \g(\STAs{I}{\top}) \sqcap \swrlG{\tmWR{D}, \tmWR{I}} \sqcap \swrlG{\tmWR{D}, \tmWR{I}}\notag\\
& \rightharpoonup \hasSt{\onto{R}, \top} \sqcap \hasTm{\onto{R}, \tmWR{D}} 
\end{align}
Finally, the overall model in \autoref{eq:m-exF} to detect the activity involving taking the objects, using them for an interval of time $\delta$ and then placing them back in the appropriate location, can be written as:
\begin{align}
\label{eq:SWRLmodel}
&\g(\STAs{T}{\top}) \sqcap \g(\STAs{R}{\top}) \sqcap \swrlA{\text{?}\bar{t},\tmWR{T},\delta} \sqcap \swrlG{\tmWR{R}, \text{?}\bar{t}} \notag\\
& \rightharpoonup \hasSt{\onto{A}, \top} \sqcap \hasTm{\onto{A}, \tmWR{R}}
\end{align}
If such a rule is evaluated as $\top$, then we say that the model is \textit{satisfied}. 
It is noteworthy that, by construction of \g(?$\STA{X}$), a generic variable \STA{X} is allowed \emph{iff} it is constrained to be an instance representing a \onto{Statement} (\autoref{eq:sem-st}). 
In this case, such an instance may also be inferred through semantic reasoning, e.g., to infer that a specific sensory data sample ($D$) has been generated from the \onto{kitchenCabinet} and that there exists a presence sensor statement (e.g., \STAs{I}{\top} or \STAs{O}{\bot}) for that specific cabinet. 

\subsection{Limitations of the Grounding and Reasoning Processes}
\label{sec:owl-limitation}

While state-of-the-art reasoning approaches over the OWL framework allow for building models and contextualizing their semantics, they also assume i) an \textit{open world} and ii) monotonic reasoning.
The first assumption limits the expressiveness of a model, whereas the latter imposes that instances be not accessed during reasoning, and that the reasoner itself cannot create new statements, but it can only manipulate existing ones.

For instance, due to the open world assumption, it is not possible to semantically infer the maximum, or minimum, element of a set, since it could be infinitely large but the system may represent only part of it. Specifically, in the example introduced above, to deal with this issue the \emph{taken} model has been defined using two different rules (\autoref{eq:sw-exT}) and this solution is clearly unsuitable for a general-purpose framework, both in terms of expressiveness and performance. 
More suitable formalisms can effectively solve this problem through the implementation of an algorithm iterating over known values.
Similarly, we could limit the effects of the monotonic reasoning assumption, versus the required number of reasoning tasks, through other languages (e.g., an extra-logic language like Java) to perform statement creation or removal within reasoning tasks.
However, in the reasoning process, data access would remain unavailable.
In order to overcome this issue, Arianna adopts the concept of external \emph{procedures}, with the aim of performing a flexible evolution of semantic data through models evaluation and external changes to the ontology.
Unfortunately, such an approach may lead to issues related to data synchronization and consistency during computation, as well as to the efficient evaluation of models in a real scenario.
Indeed, since the reasoner cannot give any priority to the SWRL rule composing a model with $n$ sequential rules, in the worst case its full evaluation occurs at the $n$-th reasoning step.
Moreover, having several SWRL rules based models managed by a single reasoner is a very expensive task, since the reasoning computational complexity is exponential with respect to the complexity of the ontology. 
Therefore, Arianna distributes the overall semantics within a network of smaller ontologies, each one with an associated reasoner.
On the one hand, a number of procedures are in charge of dynamically connecting such sub-structures with the purpose of overcoming the expressiveness limitations outlined above.
On the other hand, such a design allows for maintaining a distributed semantic structure aimed at minimizing the computational load of each reasoner, with improved performance and increased modularity. 

\section{A Distributed Network of Ontologies}
\label{sec:MON}

\subsection{Overview}
\label{sec:overview}

Arianna is composed of two types of modules:
i) a node contains an ontology describing a fragment of the context as well as its associated reasoner, and
ii) a procedure manipulates system beliefs and may also activate sensing and action algorithms.
We consider the system \textit{knowledge base} as the collection of all the nodes sharing the same statements representation (i.e., instances of the \onto{Statement} class), for which all procedures read, write, create and delete statements in a node, as well as synchronize their reasoners.
Such capabilities can be used to create a semantic network through on-line ontology manipulations. 

It is noteworthy that this structure allows for distributing the computational load associated with the overall reasoning process among consistent semantic structures with different levels of detail.
For instance, let us consider the problem of detecting a person \textit{having breakfast} based on the outcome of the previous model for items usage (\autoref{eq:SWRLmodel}) and a time representation involving the concept of \textit{morning}. 
Similarly to the previous case, an ontology representing the semantics of items taken from and released in a cabinet, as well as the related rule-based models and a time interval classification, is enough to ground such a model.
However, the resulting belief may have different semantics from the ones used in the ontology which the model is grounded on.
Such a belief could be more informative in other representations, for instance to control the \textit{meal regularity} over the current monthly time span. 
In those cases, the framework can accommodate different ontologies, which must cope with different reasoning constraints (e.g., executed many times during morning hours in one context, or once a month in the other) and semantics (e.g., use of raw sensory data, or daily reports).
This mechanism tends to improve the system's expressibility, modularity and computational performance due to representations with different levels of detail (e.g., reasoning about item positions, or long-term healthy habits).  

\subsection{Semantic Procedures}
\label{sec:procedure}

We define an external procedure as a function characterized by 
i) a set of instructions implementing an algorithm,
ii) the access to the knowledge base to obtain inputs and generate outputs and
iii) a number of activation conditions.
In particular, we consider as allowed inputs and outputs only those which scope lies within the \onto{Statement} domain, accessible through the names of the stored semantics, or to be stored in an ontology of the whole knowledge base.
This forces each procedure to share a common, distributed data structure with the main purpose of updating its state and making the system's set of beliefs evolving over time. 

During such state update or evolution, a procedure may need specific semantic requirements to work effectively and to produce consistent data.
This is represented by a set of activation conditions, defined as the Boolean matching between the system states and required statements classification.
For instance, in order to react to the fact that an object has been taken from a cabinet but not subsequently placed in its original position, a procedure may implement alerts for a user about the misplacement, which requires the object's name as input and creates the semantic description of such an alert as an output.
In this example, the activation condition would be that statements $\STA{T}$, describing whether an object has been taken before a certain time has passed, and $\STA{R}$, related to object release, are available for computing if the model $\STAs{T}{\top} + \delta < \STAs{R}{\bot}$ is satisfied, i.e., that the object has been taken and not released after a certain time interval.

\subsection{Semantic Events}
\label{sec:events}

In order to orchestrate all the procedures, the system must check whether all the specified activation conditions hold in the knowledge base, i.e., checking for the occurrence of a number of specific events over time.
We define an \emph{event} as a function returning a Boolean (\dom{B}) value based on conjunctions of classifications ($\top$) or lack of classifications ($\bot$) in specific statements of activation conditions, e.g., 
$\exists\{\onto{ObjectTaken}(\onto{T}), \onto{ObjectReleased}(\onto{R})\} \sqsubseteq \onto{Statement}$. 

The events evaluation policy is a crucial factor in terms of data consistency and performance, since events are designed to be the basic mechanism to activate reasoning procedures about models and to contextualize data.
Indeed, each activation condition may be checked with different approaches, based on the specific application semantics.
This motivates the implementation of a procedure triggering event as the collection of its activation conditions, which are treated as independent Boolean values.
Such a design choice has been done also having modularity purposes in mind, since activation conditions can be shared among events of different procedures.
For instance, the activation condition of \emph{being close to the table} can be used to describe either the event of \emph{sitting at the table} or to trigger a procedure to evaluate a \emph{having dinner} belief, or to describe a \emph{visiting the leaving room} event, aimed at triggering a procedure for \emph{cleaning} beliefs on the ontology.
In Arianna, we assume that activation conditions are updated with a specific frequency, and when their states are changed, the corresponding affected events are recomputed.
Nevertheless, other approaches could be used, for instance based on publish/subscribe policies, as well as on different heuristics, but those approaches are not supported directly by the OWL formalism.

Typically, since event evaluation is a time-critical operation, the corresponding algorithm should be carefully designed not to decrease system's performance.
For this reason, we do not perform OWL-related reasoning during the evaluation of event conditions.
In fact, we perform \textit{ad hoc} reasoning, as described in the next Section. 
The employed approach strongly reduces computational time, at the price of a reduced generality of the semantics associated with an event (i.e., the evaluating algorithm, although parameterized, is hard coded), through the access of ontology contents.
Such an operation can be delayed only when the reasoner (when remotely called) is updating the ontology, given the monotonic reasoning assumption. 
This explains why activation conditions should be much simpler than the model, in order to access as less ontological axioms as possible. Therefore, events do not check the state and time of the statements as well as their classification through statement operations; instead, they only check whether a statement exists in a given class (i.e., within a given abstract context).

Through such a combination of procedure's activation conditions, we define events not necessarily specified in time.
For instance, in the example introduced above, we fix the minimum recognition delay ($\delta$), but we do not constrain an alarm event for the object misplacement case.
Instead, the system is expected to automatically detect when the specified event occurs in the knowledge base (i.e., an activity is likely to being carried out), by checking its state over time against the activation conditions.
In practice, with a given frequency, the system accesses the ontologies to evaluate an event through condition changes: when such an evaluation is $\top$, the corresponding procedure is called and it is expected to modify a number of beliefs.

Finally, for what procedure synchronization is concerned, it is possible to introduce a statement, which is managed by a given procedure and used as activation condition for another procedure, acting as a sort of semaphore.
For more sophisticated synchronization policies, a specific semantic structure could be used as well, to consistently ground a scheduling behavior, for instanced as done in \cite{SeART}. 

\subsection{The Core Semantics of the Ontology Network}
\label{sec:core}

In our framework, as anticipated above, we introduce two types of modules, namely nodes and procedures, which can be connected together to flexibly generate a network representing the domain of interest and to reason about it, as shown in \autoref{fig:net}.
Nodes are collection of DL-based axioms evaluated by an independent OWL reasoner (which can be updated from a procedure), while the knowledge base is considered as the union of all the ontologies and inferred axioms (i.e., all the nodes).
Nodes are linked with each other through procedures, which react to semantic states (i.e., events) in order to access the knowledge base and to generate new statements, with the purpose of activating new events and updating the knowledge base accordingly.

In order to manage such a network, we designed an upper ontology describing the semantics associated with network modules by means of OWL individuals and properties.
Specifically, a \onto{w} individual is classified as a \onto{Procedure}, and therefore can be run and access the knowledge base, if it specifies: 
\begin{eqnarray}\label{eq:indP}
\label{eq:procedure}
\onto{w} \sqsubseteq_{=1}& \onto{hasImplementation}(p\in\dom{S}) \sqcap_{\geqslant 1}  \onto{requiresEvent}(\onto{e}\sqsubseteq_{\exists}\onto{Event}) \nonumber \\
&\doteq \onto{Procedure} 
\end{eqnarray}
where the \onto{hasImplementation} property indicates the unique \emph{string} identifier of the corresponding algorithmic implementation, based on Java reflection \cite{JavaReflection}.
Then, \onto{e} identifies an instance such as:
\begin{eqnarray}
\onto{e} &\sqsubseteq_{\geqslant 1}  & \onto{hasCondition}(\onto{c}\sqsubseteq_{\exists}\onto{Condition}) \doteq \onto{Event}
\end{eqnarray} 
which specifies a collection of activation conditions defined as:
\begin{eqnarray}
\label{eq:condition}
\onto{c}\sqsubseteq_{=1}& \onto{checksStatement}( x \in \dom{S}) \sqcap_{\geqslant 1}  \onto{withFrequency}( n \in \dom{N}) \nonumber\\
&\sqcap_{\geqslant 1}  \onto{hasState}( z \in \dom{B}) \doteq \onto{Condition}             
\end{eqnarray}
where $x$ indicates the desired statement (\onto{X}) and contextualization class, for instance $\onto{ObjectTaken}\sqsubseteq\onto{Statement}$, through a text string (i.e., ``\onto{ObjectTaken}(\onto{X})''), while $n$ describes the procedure execution frequency using an integer number and $z$ indicates whether the condition holds (initialized to $\bot$). 

Based on this representation, we introduce the \onto{TEvent} class through an \onto{Event} having all the \onto{Conditions} with a true state, such as:
\begin{align}
\onto{TEvent} &\sqsubseteq_{\forall} \onto{Event}.\onto{hasCondition}(\onto{TCondition})\label{eq:Tevent}\\
\onto{TCondition} &\sqsubseteq_{\exists} \onto{Condition}.\onto{hasState}(\top)
\end{align}
Then, this ontology classifies the $\onto{Runnable}\sqsubseteq\onto{Procedure}$ to be activated via an \textit{ad hoc} implementation of the rule:
\begin{equation}
\label{eq:runnable}
\onto{Procedure}(\text{?}\onto{w}) \sqcap \onto{TEvent}(\text{?}\onto{e}) \sqcap \onto{requiresEvent}(\text{?}\onto{w},\text{?}\onto{e}) \rightharpoonup \onto{Runnable}(\text{?}\onto{w})
\end{equation}
and it launches its implementation defined in \autoref{eq:procedure}.
It is noteworthy that the above rule represents the fact that a procedure with more events is triggered only if at least one of them holds true.
Instead, \autoref{eq:Tevent} defines that an event is $\top$ \emph{iff} all its conditions hold true.
Through these specifications it is possible to design all \emph{and}/\emph{or} logical operations between conditions in order to run a procedure.

Similarly, we describe also the semantics associated with nodes in the network, which is specified using:
\begin{equation}
\label{eq:node}
\onto{o} \sqsubseteq_{= 1} \onto{hasIRI}(d\in\dom{S})\sqcap_{\geqslant 1}\onto{hasReasoner}(r\in\dom{S}) \doteq \onto{Node}
\end{equation}
where the IRI is the standard ontology unique address, while the \onto{hasReasoner} property identifies the unique name of its reasoner.
Based on this classification and the Multi-Ontology Reference (aMOR)\footnote{Available at: \url{https://github.com/EmaroLab/multi_ontology_reference}}\cite{ARMOR} API, which allows for using many reasoners and for interacting with their ontology by name, this mechanism assures that each procedure can access any ontology in the knowledge base, by accessing the properties of all \onto{Node} instances.

The system's upper ontology can be reduced to a special \onto{Procedure} that manages all the characteristics of all other procedures, as discussed in Section \ref{sec:procedure}.
Specifically, such a procedure is based on a special \onto{Event} with two activation conditions:
the first checks whether a new activation \onto{Condition} belonging to another procedure exists, and if it is true, it retrieves its statement and frequency (\autoref{eq:condition}) to be forwarded to a separate periodic thread aimed at updating the \onto{hasState} property of 
$\onto{c}\sqsubseteq\onto{NewCondition}\sqsubseteq\onto{Condition}$; 
the second occurs in order to stop such an evaluation thread when
$\onto{c}\sqsubseteq\onto{OldCondition}\sqsubseteq\onto{Condition}$, disjointed from \onto{NewCondition}, is inferred.
Then, for both activation conditions, \onto{c} is reclassified as a simple $\onto{Condition}$.
The execution of the periodic thread is designed to re-evaluate all the \onto{Event} instances using such a condition, if \onto{c}.\onto{hasState} changes truth value.
If this leads \onto{e} to have all $\top$ conditions, such a thread is in charge also to trigger the procedure computation through \onto{Runnable} reasoning (i.e., it implements \autoref{eq:runnable}). The activated procedure is in charge of managing its classification in the upper ontology during all operations.

Such an architecture allows for dynamically maintaining a pool of human activity recognition modules, through their \onto{Node}, \onto{Procedure}, \onto{Event} and activation \onto{Condition} instances.
Therefore, it becomes possible to extend the semantics associated with the upper ontology to evaluate only a part of the provided models based, for instance, on temporal classification, or types of available sensors.

\section{Experimental Evaluation}
\label{sec:experiments}

\begin{figure}[!t]
\centering
\begin{center}\scalebox{.8}{\input{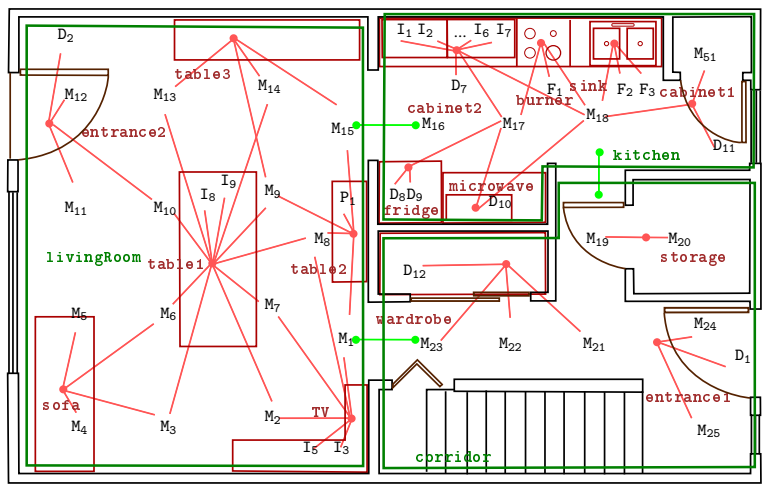}}\end{center}
\caption{The topology of the smart home of the CASAS dataset. Instances of \onto{Rooms} (in green) and \onto{Furniture} (in red) are indicated with points linked to \onto{sensors} through the property \onto{isNearTo}, while the \onto{isLocatedIn} property is visualized through their locations.}
\label{fig:casasTopology}
\end{figure}
\subsection{The Used Dataset}
\label{sec:dataset}

The overall behavior of the Arianna inference engine has been tested using the WSU CASAS Dataset\footnote{Available at \url{ailab.wsu.edu/casas/datasets/adlinterweave.zip}}.
The dataset is described in \cite{casas} and contains human activity data collected while performing experiments in an apartment equipped with several distributed sensors, as shown in \autoref{fig:casasTopology}.
Here, we assume that only \textit{one} person is performing daily activities.
The data originating from the available sensors are described in the \onto{Statement} domain as: the presence of the assisted person (\STA{M}), the presence of items (\STA{I}), the position of the phone handset (\STA{P}), the state of doors or shutters (\STA{D}) and the state of the gas or water flows (\STA{F}).
For the sake of this case study, we discard temperature and brightness information since we do not use them in any activity recognition model discussed here. 
Within this environment, the experimental protocol foresees that the person performs eight activities:
\begin{enumerate*}[label=(\STA{A_\arabic*})]
  \item fill medication dispenser,
  \item watch DVD,
  \item water plants,
  \item converse on the phone,
  \item write a card,
  \item prepare meal,
  \item sweep and clean, and
  \item select an outfit.
\end{enumerate*}
Twenty participants followed such a script (of approximately a quarter of an hour) twice, in different days.
The first time they are asked to perform the set of activities in a sequential way, while the second time they are left free to perform the activities in any order and with possible interruptions. 
We refer to the first case as \textit{sequential} dataset, and the second as \textit{interwoven} dataset.
During those experiments, the information generated by the sensors has been stored and distributed along with a temporal stamp and activity labels, which we use to validate our framework.


\subsection{Setting up the Ontology Network}
\label{sec:setUp}

\begin{figure*}
\centering
\begin{center}\scalebox{.75}{\input{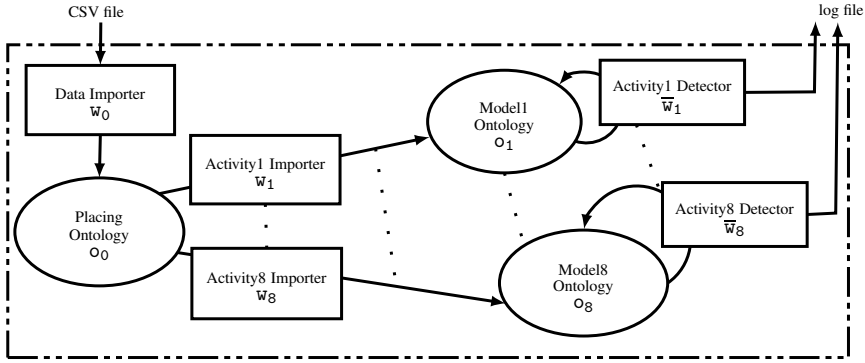}}\end{center}
\caption{The ontology network setup for the CASAS dataset semantically specified in the system's core. Rectangles identify a \onto{Procedure} and circles stand for \onto{Nodes} (i.e., ontologies), while arrows show statements evolution for the recognition of the considered activities.}
\label{fig:net}
\end{figure*}

\autoref{fig:net} shows the Ontology Network configuration designed to model the CASAS scenario.
We designed an upper ontology \PlaceOnto{} aimed at spatially contextualizing sensor-related statements.
Based on such descriptions, we specified the used events with a procedure for each activity, referred to as \TImporter{i}, which computation is related to the evaluation of the human activity recognition model represented in \ModelOnto{i}.
Just for the sake of this simulation, when a model is satisfied, the computation of the related \TDetecter{i} procedure is triggered to notify activity recognition.
It is noteworthy that the fact that such a network configuration consists of the same structure for each activity shows the system modularity in describing models with different levels of detail.

A number of nodes and procedures have been implemented.
Two classes of \onto{Nodes} are present: 
\begin{enumerate*}[label=(\emph{\roman*})]
\item the \emph{placing ontology} \PlaceOnto{}, which contextualizes sensory data with respect to the topology of the environment, and infers the assisted person's location, and  
\item a set of \emph{model ontologies} $\ModelOnto{1}, \ldots, \ModelOnto{8}$ for representing each activity recognition models. 
\end{enumerate*}
As far as the implemented procedures are concerned: 
\begin{enumerate*}[label=(\emph{\roman*})]
\item the \emph{data importer} \DImporter{} simulates sensory data streaming in \PlaceOnto{} and updates its beliefs;
\item the \emph{activity importers} $\TImporter{1}, \ldots, \TImporter{8}$ collect spatially contextualized statements from \PlaceOnto{} and \textit{move} them in ontologies describing activity recognition models (\ModelOnto{i}), then they update their reasoners, which generate a specific $\top$ statement if the associated activity is recognized;
\item the set of \emph{activity detectors} $\TDetecter{1}, \ldots, \TDetecter{8}$ listen, in the related \ModelOnto{i} nodes for such a recognition event, with the purpose of notifying the recognition and resetting the model for further evaluations.
\end{enumerate*}

While model ontologies share common semantic descriptions, e.g., the time representation (although different \emph{resolutions} can be specified), and the format of outcome statements, which is $\top$ when the activity is recognized, the placing ontology implements a dedicated semantics describing the environment, and it is used as a middle abstract layer between raw sensory data and model beliefs.
The \PlaceOnto{} ontology describes the topology shown in \autoref{fig:casasTopology} in terms of \onto{Room}, \onto{Furniture} and \onto{Sensors} (i.e., classes), as well as \onto{isNearTo} and \onto{isLocatedIn} (i.e., relations). 
Instances representing all the components in the topology (i.e., sensors and locations in \autoref{fig:casasTopology}) belong to one of those classes, and are also related to each other through such relations.
This makes the reasoner infer the position of the assisted person with respect to a location that is semantically defined (e.g., \onto{Cabinet2} if \STAs{D_7}{\top}).
Therefore, adopting such a contextualization, we exploit the ability of all the sensors (not only of motion detectors) to store in the \PlaceOnto{} the assisted person's location.

Such a representation is populated by the data streaming and the reasoning updates performed by \DImporter{}, which performs a memory-free introduction of statements in the ontology.
This forces \PlaceOnto{} not to reason temporally, namely, samples related to the same sensor are updated over time.
The temporal statement definition is preserved for further propagations, but the reasoner of \PlaceOnto{} does not consider it.
This is meant at upper bounding the number of axioms (i.e., the ontology \textit{complexity}) in the placing ontology, with the purpose of efficiently managing raw data against soft real-time constraints.
Indeed, \PlaceOnto{} is the representation with the highest reasoning rates and the belief associated with the assisted person's position is fundamental for most of the activity recognition models.
As a consequence, reasoning in the placing ontology is a crucial part for the overall system performance, the robustness of such a design be assured if no activity models and sensors require a temporal resolution smaller than the reasoning time in \PlaceOnto{}.

As far as the computation of the \TImporter{i} and \TDetecter{i} procedures is concerned, we replicated a common pattern to implement in the network a generic activity recognition package (\autoref{fig:net}). 
The pattern is centered in the \ModelOnto{i} node, which represents the semantic model for the $i$-th activity. 
This is the only module that changes considerably among different activity network branches. 
As an instance, the third activity model requires an ontology $\ModelOnto{3}$ that: 
\begin{enumerate*}[label=(\emph{\roman*})]
\item describes the belief that a watering can should be filled, as well as
\item uses the statement and 
\item its spatial semantics, both shared with \PlaceOnto{} (e.g., that the environment has a door sensor in a certain room with a certain state at a given time) and finally,    
\item knows the location where the watering can should be located.
\end{enumerate*} 

The activity importer \TImporter{i} represents the computation performed to evaluate such \ModelOnto{i} model through instance checking.
Its activation conditions are such to identify when the person is in the location where the activity is supposed to be performed, while the triggered computation is aimed at propagating, from the placing ontology, relevant statements for the \ModelOnto{i} model, and then to update its reasoner.
While reasoning on the $i$-th activity model, which corresponds to the evaluation of a set of SWRL rules, an activity recognition belief can be generated (i.e., $\STAs{A_i}{\top}$).
This triggers the corresponding activity detector \TDetecter{i}, since its activation condition is such to listen for that specific statement classification (e.g., \onto{Recognized}($\onto{A}_\onto{i}$), which did not exist before, therefore the condition was $\bot$).
When this occurs, the $\TDetecter{i}$ procedure is also in charge of resetting the \ModelOnto{i} representation by removing all statements, which is a crucial feature for continuous reasoning performance.

\subsection{Statements, Models and Events}
\label{sec:model}

\begin{figure}
\begin{center}\hspace*{-.5cm}
\begin{center}\scalebox{1}{\input{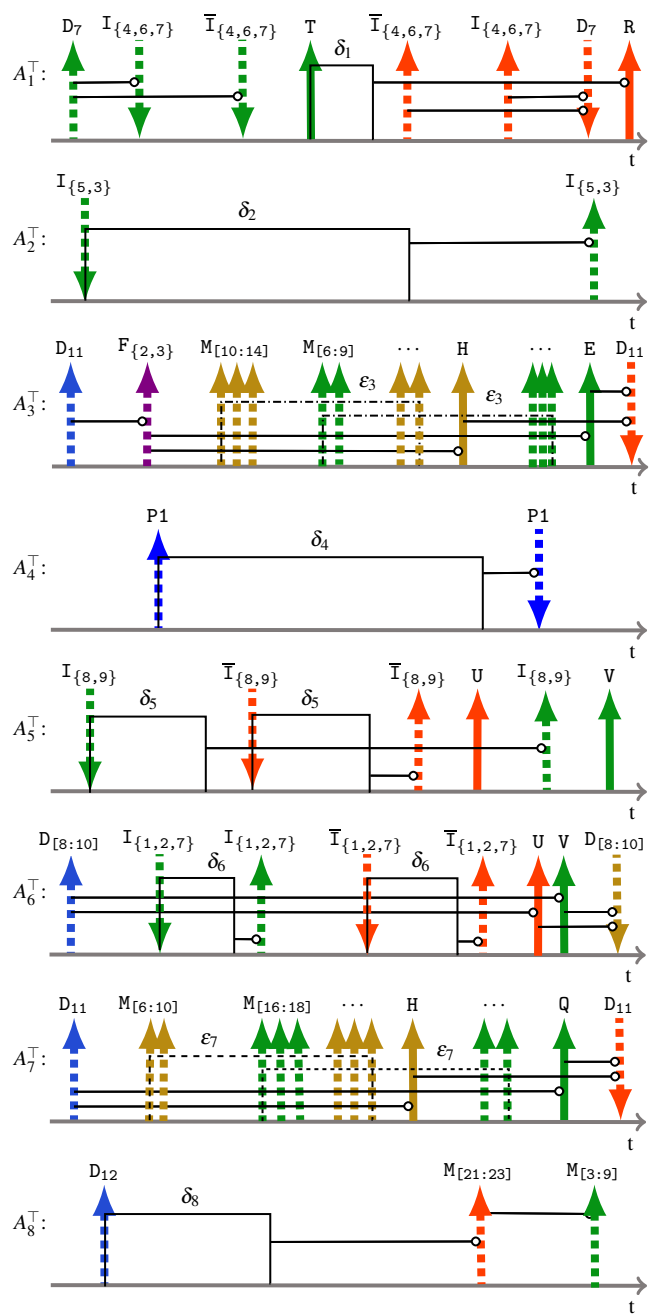}}\end{center}
\caption{Activity recognition models for the CASAS dataset. Required statements are shown as vertical lines where: color represents sub-rules and dashed lines the spatial level of representation details. Sensors are indicated with a unique index value (i.e., curly brackets) or with a list (i.e., squared brackets), while temporal restrictions are shown as black lines.}
\label{fig:models}
\end{center}
\end{figure}

\autoref{fig:models} shows a graphical representation of the fluent models introduced in Section \ref{sec:arianna_computational_framework}, where statements are shown as vertical arrows pointing upwards to indicate a $\top$ state and downwards for $\bot$. 
These statements are created directly from sensors.
We indicate their type and index, or a range of indexes, where we divide sets, denoted by curly brackets, from lists, denoted by squared brackets.
If a set is used, the same incoming index is considered for all the same statements of the model.
For instance, in model \STA{A_1}, if \STA{I_6} and \STA{\bar{I}_4} are satisfied in the left part of the graph, then only the same indexes are considered on the right.
Instead, if a list is used, any possible statements generated by those sensors are processed by the model. 
It is noteworthy that the sensors indicated in each graph identify the statements that \TImporter{i} should propagate when it is triggered, i.e., the relevant information for the model.

Statements are annotated along a relative $x$-axis, in order to restrict their temporal relations through black lines ending with a circle.
Given two statements and their generation times, respectively where the line starts \tm{J} and ends \tm{Y}, the black line represents that $\tm{J}+\delta<\tm{Y}$ must hold true for the model to be satisfied.
Thus, each graph in \autoref{fig:models} indicates the rule restrictions of the activity recognition models, expressed in SWRL rues as in \autoref{eq:SWRLmodel}, namely if they hold in the relative \ModelOnto{i} ontology, the system generates an $\STAs{A_i}{\top}$ belief for specifying that the $i$-th activity has been recognized.

In \autoref{fig:models}, we used the same color to highlight the statements used in sub-rules for computing intermediate beliefs required by the models.
For instance, the first graph represents the model of \emph{filling a dispenser}, grounded in \autoref{eq:SWRLmodel} (where we substitute \STA{O} with \STA{\bar{I}}), which can be considered as the problem to take (\STA{T}) and release (\STA{R}) some items, shown in green and red (Equations \ref{eq:m-exT} and \ref{eq:m-exR}), respectively.
From the graphical representation of the \STA{A_1} model, it is possible to appreciate that the \STA{D_7} cabinet door is open before that item absence statements \STA{I} and \STA{\bar{I}} hold true, which makes the green rule satisfied (i.e., the reasoner infers the \STAs{T}{\top} statement). 
Then, the overall model is satisfied when this happens before than $\delta_1$ time units with respect to the \STAs{R}{\top} statement, which is similarly generated via the red rule.
 
It is worth noting that no theoretical limitations in the number of levels of detail for a model are assumed.
However, in our case study, we never used models with more than one intermediate level, i.e., the spatial contextualization.
Therefore, we highlighted with dashed statements the information coming from the placing ontology, while with filled statements the beliefs generated by the model in the related \ModelOnto{i} ontology, where we did not overlay the filled line with the last dash statement of the rule, as defined in Equations \ref{eq:m-exT} and \ref{eq:m-exR}.

In a model, it is possible to specify not only the type of sensors providing relevant statements, but also the activity importer events for an efficient model evaluation.
For instance, we designed \TImporter{1} using a single event as an activation condition concerning the presence of the assisted person \emph{close} to \onto{cabinet2} (\autoref{fig:casasTopology}).
With this design, the fluent model in charge of evaluating \STA{A_1} can use only the information necessary for its computations.

Let us now briefly introduce the semantics of all other models in \autoref{fig:models}, as well as their activation events. 
\begin{enumerate}[label=(\STA{A_\arabic*})]\setcounter{enumi}{1}
\item The \textit{watch DVD} model is similar to \STA{A_1} but assumes the availability of a presence statement regarding possible DVD locations (\STA{I}); 
based on it, the \STA{A_2} activity is recognized if the item is missing from its position for at least a delay of $\delta_2$ time units; for this activity we consider an event based on the condition of the person being in the \onto{livingRoom}.
\item The \textit{water plants} model is more complicated and requires the beliefs that: the door (\STA{D}) of the room where the watering can is supposed to be located is opened and later the water sensor becomes active (\STA{F}); then the model counts the amount of time spent near \onto{table3} (yellow statements) and \onto{table1} (green statements), where some plants are known to be located, and when this becomes longer than a threshold $\epsilon_3$ the \STA{H} and \STA{E} statements are generated, respectively; finally, the model is satisfied only if $\STAs{D}{\top} < \STAs{F}{\top} < \STAs{H}{\top} < \STAs{D}{\bot}$ and $\STAs{D}{\top} < \STAs{F}{\top} < \STAs{E}{\top} < \STAs{D}{\bot}$, namely when the watering can is also filled and replaced in its location, associated with the closure of the door after having visited the plants location.
Such model design requires that its computation is carried out for more than one location, therefore we set the \TImporter{i} to have three events, each one based on a unique condition for the assisted person to be located: \emph{near} to \onto{cabinet1} or \onto{sink}, or \emph{in} \onto{livingRoom}.
\item To model the \textit{converse on the phone} model we consider the usage of the phone handset (\STA{P}) for at least $\delta_4$ time units. 
In this case, the specified event for the activity importer is simply near to the \onto{table2} furniture.
\item The \textit{write a card} model is similar to \STA{A_1}, since it describes the usage (i.e., taken and released after $\delta_5$ time units) of two items, namely \STA{I} and \STA{\bar{I}}; however, in this case, no door statements are considered. 
Similarly set is also the activation condition, which is based on the occupancy of the \onto{table1} semantic area.
\item The \textit{prepare meal} model is based on the usage (based on $\delta_6$, similarly to \STA{A_5}) of two kitchen items \STA{I} and \STA{\bar{I}}, taken from a cabinet \STA{D}, which should be closed at the end of the activity.
In this case, we consider as a triggering event the fact that the assisted person is in the \onto{kitchen}.
\item To model the \textit{sweep and clean} activity, we used the same approach of \STA{A_3} by detecting that the cleaning items have been taken from a cabinet \STA{D} and that the person spends some time ($\epsilon_7$) in the living room (yellow statements) and in the kitchen (in green).
Similarly to what has been done before, configured events for this models are two, each one with a unique condition (i.e., in \emph{or}): one if the person is in the \onto{livingRoom}, while the other if the person is in the \onto{kitchen}.
\item Finally, the \textit{select an outfit} model is designed to assess whether the person remains in the corridor, after the wardrobe has been opened, for at least $\delta_8$ time units, before the person moves where he/she is supposed to leave the chosen outfit.
In this case, we did not rely on the closure of the wardrobe due to many false-positive samples in the dataset.
Similarly to the previous case, we set three events assessing that the person is: in the \onto{corridor}, or near the \onto{sofa} or \onto{table1}.
\end{enumerate}

\subsection{Results}
Experiments have been conducted using the setup described in the previous Section and with the dataset introduced in Section \ref{sec:dataset}. 
The simulation is performed with a real-time data stream in order to evaluate the overall system behavior against the provided activity labels, in terms of recognition rates and time performance. 
For both the datasets, sequential and interwoven, we randomly concatenate the twenty experiments with a delay of three minutes  to simulate a scenario where people enter in the apartment one after the other, while contextualized data cleaning (i.e., deleting axioms in \ModelOnto{i}) is performed when the $i$-th activity is recognized or no new statements are generated by \DImporter{}.
All results have been generated on a single machine with processor Intel\textsuperscript{\tiny\textregistered} Core{\tiny\textsuperscript{\texttrademark}} $i5$-$460M$ $2.53GHz$ and $4GB$ of memory. 

\begin{figure}[]
	\centering
	\newlength\figureheight 
	\newlength\figurewidth 
	\setlength\figureheight{3cm} 
	\setlength\figurewidth{.85\columnwidth} 
	{\hspace*{-0.6cm} \input{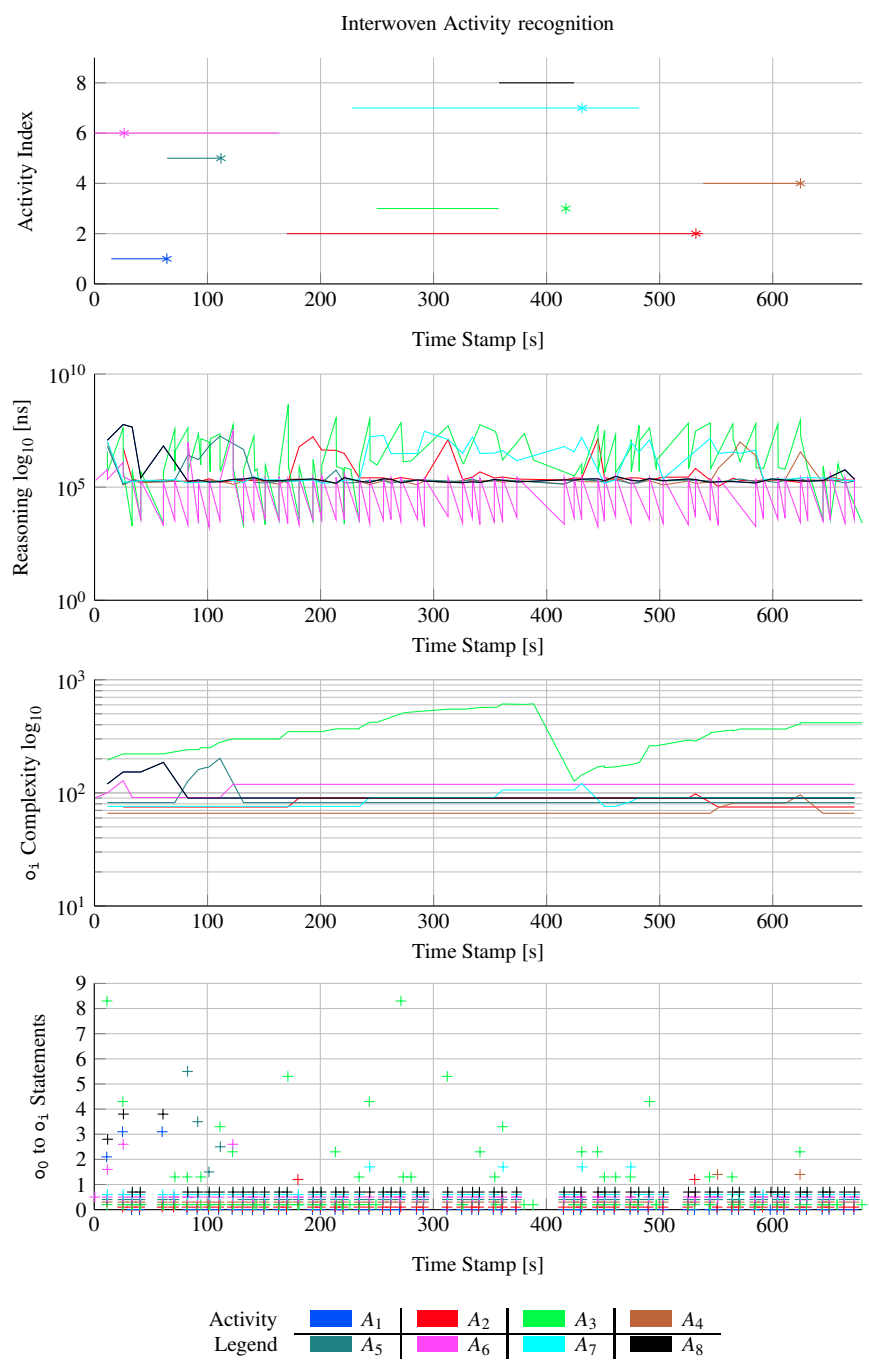}\\}
	\vspace{.3cm}
     \begin{tabular}{cc|c|c|c}
     \multirow{2}{*}{\begin{tabular}[c]{@{}r@{}}Activity\\ Legend\end{tabular}} 
          & \coloredBox{Leg1}~\STA{A_1} & \coloredBox{Leg2}~\STA{A_2}  & \coloredBox{Leg3}~\STA{A_3}  & \coloredBox{Leg4}~\STA{A_4} 
          \\ \cline{2-5} 
          & \coloredBox{Leg5}~\STA{A_5} & \coloredBox{Leg6}~\STA{A_6}  & \coloredBox{Leg7}~\STA{A_7}  & \coloredBox{Leg8}~\STA{A_8} 
     \end{tabular}
	\caption{The system behaviour during the simulation with the 30th volunteer for all the interwoven activities in the CASAS dataset. The top graph shows the experiment and the associated recognitions, followed by the two intermediate graphs showing reasoning performance, and the bottom graph with network events and statements propagated from the placing to the relevant activity ontology.}
\label{fig:subPlot}
\end{figure}

\autoref{fig:subPlot} shows the main behavior of the framework while all activities are performed in an interwoven manner.
The Figure shows four graphs with the same experimental time stamp ($x$-axis) and a color map, indicating all activity indexes from $\STA{A_1}$ to $\STA{A_8}$. 
The first graph describes dataset labels using a horizontal line (i.e., the time spent in performing an activity) while the `\textasteriskcentered' symbol indicates the time instant when the system notifies such activity has been executed.
The $y$-axis represents the $i$-th activity index (according to Section \ref{sec:dataset}), and it is aimed at showing the activity ground truth against the recognition of the 30th volunteer, as well as at providing the temporal performance below with a reference.
The second and the third graphs represent, with a logarithmic scale, the reasoning performance for each model (i.e., all \ModelOnto{i}), and specifically the computation time, in nanoseconds, and the ontology complexity (i.e., the number of axioms). 
The fourth graph shows the number of statements propagated from the placing ontology \PlaceOnto{} to the related model ontology \ModelOnto{i} through the activity importer procedure \TImporter{i}, shown in \autoref{fig:net}, where each branch of the network is identified using a different color.
Therefore, it provides a basic mechanism to observe all reasoning behaviors with respect to the placing contextualization and the network evolution, since each point describes the activation of a model event, which triggers \TImporter{i}, thereby increasing the \ModelOnto{i} complexity and updating its reasoner.
In this visualization, we shift the points for avoiding `$+$' symbols overlapping with both axes.
However, their real Cartesian position is at the minimum integer approximation.
From such a representation, it is possible to observe that the activation of \TImporter{i} not always introduces new statements, since events can be satisfied even if in the placing ontology no novel relevant statement for the model are available.
Nevertheless, the \ModelOnto{i} reasoner is updated anyway by the activity importer in order to update temporal classification in the activity ontology.

From the first graph of \autoref{fig:subPlot}, we can have an overview of the interruptions of the activities, as well as the performance of the recognition models shown in \autoref{fig:models}, which produce different temporal behaviors. 
In particular, \STA{A_2} and \STA{A_4} have a well-contextualized end points in the models, as the release of a single item; therefore, their recognition well fits at the end of the labeled performing line. 
Similarly, the recognition of the activities \STA{A_1} and \STA{A_5} follows the same pattern, even if they rely on a second statement layer related to two objects, which must be taken and then released (i.e., \STA{T}, \STA{R} and \STA{U}, \STA{V}, respectively). 
For the activities where we classify statements based on the time spent in a particular room or area, namely \STA{A_3} and \STA{A_2}, the recognition is clearly affected by the introduced heuristics employing the $\epsilon_3$ and $\epsilon_7$ thresholds. 
Since the activity importer \TImporter{3} and \TImporter{7} have overlapping conditions (i.e., being in the kitchen) and share in the model the change of state related to the same door sensor, a $\bot$ state of such a statement recovers \STA{A_3} for not being recognized during the performance of \STA{A_7}.
Given that $\STAs{D_{11}}{\bot}$ did not satisfy the $\epsilon_3$ threshold while \STA{A_3} was performed, and such rule was satisfied during the execution of \STA{A_7}.
Moreover, for \STA{A_6}, which has a similar shape for representing a much more complex task as cooking, the system triggers the recognition earlier due to the simplicity of the model. 
Finally, \STA{A_8} was not detected since the $\delta_8$ threshold was not satisfied, namely, the person took less time than the threshold for choosing an outfit.

Through the analysis of the reasoning performance (\autoref{fig:subPlot}), it is possible to observe a correlation between the number of axioms of a model ontology (in the third graph) and the reasoning time (in the second) of the same activity representation (\ModelOnto{i}), where the initial differences between complexities describe the context complexity of each model. 
In the reasoning process, not all models infer the same amount of information, which generates the different behaviors shown in the third graph. 
Specifically, it is possible to notice the effects of \TDetecter{i}, which resets the node complexity to its initial value when an activity is recognized (i.e., cleans \ModelOnto{i}).
Furthermore, it is possible to observe that the complexity is affected by a strong non-linearity with respect to the number of propagated statements (fourth graph), which further penalizes the reasoning performance measured \textit{after} the caching algorithm of the Pellet reasoner.
For instance, \STA{A_3} is evaluated whenever the person is in the living room and, after a while, it requires to reason about many past location statements (to check for the $\epsilon$ threshold), which increases the complexity and affects the reasoning time.
We noticed a similar behavior also for \STA{A_7}, and we argue that it is due to a lack in the model contextualization, since the node is filled with many statements, which are related to other activities.
For more contextualized models, such as $\STA{A_5}$, we can notice that the complexity does not change over the entire experiment, since its evaluation is based on sporadic events in the dataset (e.g., when a particular item is absent).
This is well described in the placing ontology, and the activity importer \DImporter{} propagates few statements when the activity is performed (first and fourth graphs).

In the second graph of \autoref{fig:subPlot}, we can observe a rough periodic behavior for each activity, which is due to the \TImporter{i} frequent event checking nature triggering its computation. Specifically, when \PlaceOnto{} changes, it meets the \TImporter{i} activation conditions and it updates the \ModelOnto{i} reasoner in order to evaluate the activity model, especially for temporal statement classifications, thereby producing computational spikes depending on the event evaluation frequency and the state of the knowledge base. 
Such a reasoning behavior is applied only to a sub-contextualized ontology, with respect to its previous state, which may be very similar, therefore the reasoning time is restrained in the magnitude of $10^{-4}$ seconds. 
For instance, in \STA{A_6} the reasoning time increases in correspondence to new data propagated from \PlaceOnto{} to \ModelOnto{6}, e.g., the latter increases its complexity, the model becomes satisfied and, after a while, it is reset.
Then, other statements are generated from \PlaceOnto{} validating the events triggering \TImporter{6}, which modifies the \ModelOnto{6} in a state where the model is not satisfied.
Such procedure performs reasoning, in order to check whether \emph{at this time} the model is satisfied, but this never happens again.
Therefore, while the complexity of the ontology does not substantially change, the reasoner has still work to do for the time classification update. 

\begin{figure}
\centering
\setlength\figureheight{9cm} 
\setlength\figurewidth{.8\columnwidth} 
\input{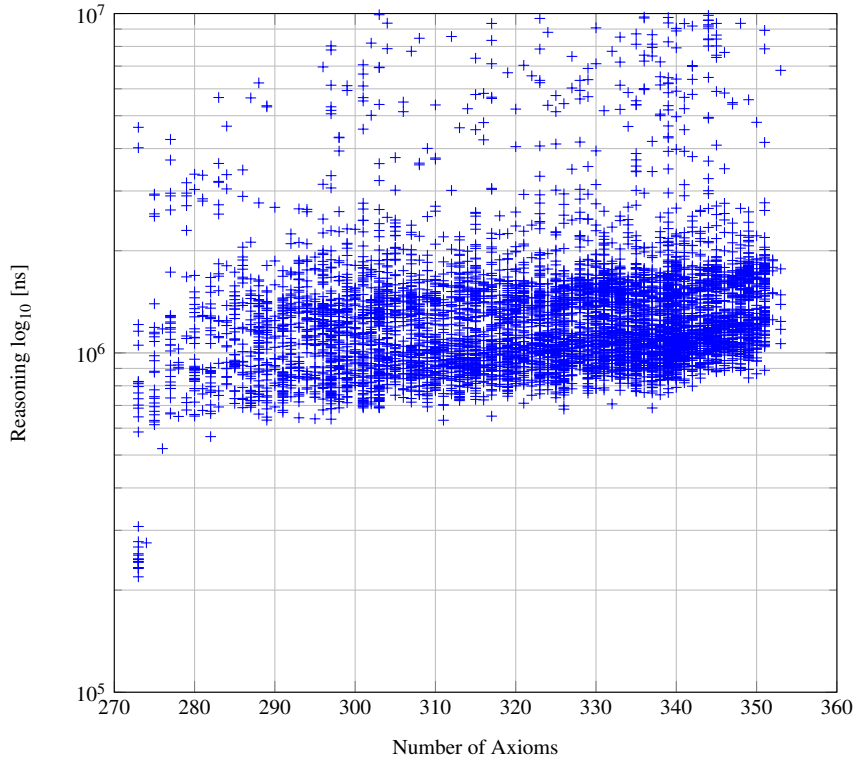}
\caption{The placing ontology (\PlaceOnto{}) reasoning time against ontology complexity measured during the simulation of the CASAS interwoven dataset.}
\label{fig:placingTime}
\end{figure}

\autoref{fig:placingTime} shows the reasoning performance of the placing ontology observed along all the interwoven dataset.
Each time new sensory data is introduced by the \DImporter{} procedure, we measure the reasoning time in nanoseconds, expressed in a logarithmic scale, and the ontology complexity. 
We can notice that the time is upper bounded of approximately the same magnitude for all \ModelOnto{i} ontologies shown in the second graph of \autoref{fig:subPlot}.
Nevertheless, its variance is smaller due to the memory-free policy of this ontology (i.e., statements are overwritten over time), which bounds the complexity to at most $353$ axioms. 
This is a crucial measure for the design of the event evaluation frequency, since \PlaceOnto{} is the ontology with the fastest reasoner update rates, and it is the base for all other models.
In particular, during reasoning time, due to the monotonic reasoning assumption, it is not possible to access nor manipulate the information, which may cause the loss of data since it is overwritten in real-time if the reasoning time is lower that the sensory data updating rate.
This is the main reason why particular attention has been devoted to the performance of \PlaceOnto{}, which well performs even if its complexity is higher than most of the other ontologies in the network. 
Such results confirm that a system composed of a unique ontology for all the models and contextualizations would not be suitable even for soft real-time purposes, since the sum of all the measured reasoning times is known to scale exponentially.

We also collected data about reasoning performance of the core procedure, which manages all event evaluations and procedures triggering in order to run a network that, for this case study, is static throughout all the experiments.
We observe that the computational time is between one or two orders of magnitude smaller than in the placing ontology, and its variance is affected by the monotonic reasoning assumption even if its computation does not rely on OWL-based reasoning, as shown in Section \ref{sec:core}. 

As far as the soft real-time performance of the overall network is concerned, we tested the system with an input data stream actually \textit{faster} than the real-world in order to stress recognition performance.
We used different increasing speed scaling factors, and we noted that activity recognition performance do not change significantly up to a speed factor of four (i.e., the input data stream is $4\times$ faster than the real case).
Such a behavior is due to the fact that the \PlaceOnto{} reasoner is able to update its contents without loosing statements crucial for models evaluation.
For similar reasons, also with smaller speed scaling factors, we noticed delays in the activity recognition instants, but this did not affect recognition since no activity ontologies employ a statement overwriting policy.

The $4\times$ speed scaling factor is useful to further address the real-time performance of the framework.
\autoref{tab:delay} shows how many times a recognition has been notified late by the system, as well as the associated worst and average values, where the best value has been considered as no delay with respect to the last activity annotation in the dataset, over the interwoven case.
In particular, it is shown that \STA{A_6} is the most delayed model, partially due to the $\epsilon_3$ heuristic threshold, in contrast with the $\epsilon_7$ threshold that seems to be well configured for this dataset.
Even simpler models can be strongly delayed, due to the data contextualization reasoning process.
For instance, \STA{A_6} is based on a simple model but since a phone call may happen during the execution of other activities, it is likely that other reasoners are in the process of verifying their models and increasing their complexity, and therefore such a load can affect the evaluation of \STA{A_6}.

\begin{table}[]
	\centering
	\begin{minipage}{0.5\linewidth}
		\centering
		\begin{tabular}{c|ccc}
			\textit{\begin{tabular}[c]{@{}c@{}}Activity\\ Index\end{tabular}} & \textit{\begin{tabular}[c]{@{}c@{}}number\\ of delays\end{tabular}} & \textit{\begin{tabular}[c]{@{}c@{}}maximum\\ delay\end{tabular}} & \textit{\begin{tabular}[c]{@{}c@{}}averaging\\ delay\end{tabular}} \\ \hline
			1				& 0/20				& $-$			& $-$	\\
			2				& 4/20				& 9.46 s			& 4.17 s	\\
			3				& 8/20 				& 17.3 s			& 9.2 s	\\
			4				& 5/20 				& 12.1 s			& 5.83 s	\\
			5				& 0/20				& $-$			& $-$	\\
			6 			     & 1/20				& 4.17 s 			& $-$	\\
			7				& 1/20				& 4.6 s 			& $-$	\\
			8				& 5/20				& 9.2 s			& 5.41 s
		\end{tabular}
	\end{minipage}
	\hfill
	\begin{minipage}{0.40\linewidth}
		\captionof{table}{Activity recognition delays measured with the CASAS dataset in a simulation performed four times faster than the real-world.}	     
		\label{tab:delay}
	\end{minipage}
\end{table}

We collected also the recognition rates shown in \autoref{tab:recognition}, which indicate an overall improvement with respect to the original results discussed jointly with the publication of the dataset \cite{casas}:
\begin{align}
\begin{tabular}{r|cccccccc}
\textit{Activity Index} & \textit{1} & \textit{2} & \textit{3} & \textit{4} & \textit{5} & \textit{6} & \textit{7} & \textit{8} \\ \hline
\textit{\%} & 65.6 & 86.2 & 28.4 & 58.9 & 82.8 & 82.6 & 88.1 & 67.3
\end{tabular}\nonumber
\end{align}
In particular, \autoref{tab:recognition} summarizes the percentage of true-positive and true-negative recognition rates obtained by using the system configuration shown in Section \ref{sec:setUp}, both for the sequential and interwoven activity performance. 
These results take in account how many times each model has been satisfied with respect to the activity annotation available in the datasets, and shows the rate of non recognized (i.e., \emph{unknown}) activities, while its symmetric column is omitted since it is empty (i.e., all $0\%$), coherently with the fact that results are obtained from a fully supervised dataset.
We noticed that the system usually does not recognize an activity rather than missclassifying it.
This is due to the data contextualization process, which tends not to be confused by the activity type.
Instead, in this case a model increases its complexity without being satisfied and misses the recognition.
Missclassifications may occur when shared statements are processed by different models, such as the presence in a room for an certain time, for instance in \STA{A_7}.

\begin{table*}[]
  \centering
  \subfloat[Sequential activity script\label{tab:sequentailRate}]{%
    \begin{tabular}{cc|ccccccccl}
			\multicolumn{2}{l|}{\multirow{2}{*}{\diagbox{MON}{Label}}} & \multicolumn{8}{c}{\textit{Activity Index}} & \multirow{2}{*}{} \\
			\multicolumn{2}{l|}{} & \textit{\textbf{1}} & \textit{\textbf{2}} & \textit{\textbf{3}} & \textit{\textbf{4}} & \textit{\textbf{5}} & \textit{\textbf{6}} & \textit{\textbf{7}} & \textit{\textbf{8}} &  \\ \cline{1-10}
			\multirow{8}{*}{\hspace{.3cm}\begin{turn}{-270}\textit{Activity Index ($y$)}\end{turn}} 
					& \textit{\textbf{1}} & 90 	& $-$ 	& $-$ 	& $-$ 	& $-$ 	& $-$ 	& $-$ 	& 
							\multicolumn{1}{c}{$-$} & \multirow{9}{*}{\hspace*{-.3cm}\begin{turn}{-270}\textit{recognition rates (\%)}\end{turn}} \\
					& \textit{\textbf{2}} & $-$ 	& 100 	& $-$ 	& $-$ 	& $-$ 	& $-$ 	& $-$ 	& 
							\multicolumn{1}{c}{$-$} &  \\
					& \textit{\textbf{3}} & $-$ 	& $-$	 & 80 	& $-$ 	& $-$ 	& $-$ 	& $-$ 	& 
							\multicolumn{1}{c}{$-$} &  \\
					 & \textit{\textbf{4}} & $-$ 	& $-$ 	& $-$ 	& 90 	& $-$	 & $-$	 & $-$ 	& 
					 		\multicolumn{1}{c}{$-$} &  \\
					& \textit{\textbf{5}} & $-$ 	& $-$ 	& $-$ 	& $-$ 	& 95 	& $-$ 	& $-$ 	&
							\multicolumn{1}{c}{$-$} &  \\
					& \textit{\textbf{6}} & $-$ 	& $-$ 	& $-$ 	& $-$ 	& $-$ 	& 75 	& 10		 & 
							\multicolumn{1}{c}{$-$} &  \\
					& \textit{\textbf{7}} & $-$ 	& $-$ 	& 15		& $-$ 	& $-$ 	& $-$ 	& 85 	& 
							\multicolumn{1}{c}{$-$} &  \\
 					& \textit{\textbf{8}} & $-$ 	& $-$	 & $-$ 	& $-$	 & $-$ 	& $-$ 	& $-$ 	& 
							\multicolumn{1}{c}{100} &  \\
		\multicolumn{2}{c|}{\textbf{unknown}} & 10 & $-$ & 5 & 10 & 5 & 25 & 5 &
							\multicolumn{1}{c}{$-$} & 
		\end{tabular}
  }\hfill%
  \subfloat[Intewoven activity script\label{tab:interwRate}]{%
    \begin{tabular}{cc|ccccccccl}
		\multicolumn{2}{l|}{\multirow{2}{*}{\diagbox{MON}{Label}}} & \multicolumn{8}{c}{\textit{Activity Index}} & \multirow{2}{*}{} \\
		\multicolumn{2}{l|}{} & \textit{\textbf{1}} & \textit{\textbf{2}} & \textit{\textbf{3}} & \textit{\textbf{4}} & \textit{\textbf{5}} & \textit{\textbf{6}} & \textit{\textbf{7}} & \textit{\textbf{8}} &  \\ \cline{1-10}
		\multirow{8}{*}{\hspace{.3cm}\begin{turn}{-270}\textit{Activity Index ($y$)}\end{turn}} 
				& \textit{\textbf{1}} & 95 	& $-$ 	& $-$ 	& $-$ 	& $-$ 	& $-$ 	& $-$ 	& 
						\multicolumn{1}{c}{$-$} & \multirow{9}{*}{\hspace*{-.3cm}\begin{turn}{-270}\textit{recognition rates (\%)}\end{turn}} \\
				& \textit{\textbf{2}} & $-$ 	& 95 	& 5   	& $-$ 	& $-$ 	& $-$ 	& 5   	& 
						\multicolumn{1}{c}{$-$} &  \\
				& \textit{\textbf{3}} & $-$ 	& $-$	& 70 	& $-$ 	& $-$ 	& $-$ 	& 5   	& 
						\multicolumn{1}{c}{$-$} &  \\
				 & \textit{\textbf{4}} & $-$ 	& $-$ 	& $-$ 	& 100 	& $-$	 & $-$	& 5   	& 
				 		\multicolumn{1}{c}{$-$} &  \\
				& \textit{\textbf{5}} & $-$ 	& $-$ 	& $-$ 	& $-$ 	& 80 	& $-$ 	& $-$ 	&
						\multicolumn{1}{c}{$-$} &  \\
				& \textit{\textbf{6}} & $-$ 	& $-$ 	& $-$ 	& $-$ 	& $-$ 	& 70 	& 5		 & 
						\multicolumn{1}{c}{$-$} &  \\
				& \textit{\textbf{7}} & $-$ 	& $-$ 	& 25		& $-$ 	& $-$ 	& $-$ 	& 80 	& 
						\multicolumn{1}{c}{$-$} &  \\
				& \textit{\textbf{8}} & $-$ 	& $-$	& $-$ 	& $-$	 & $-$ 	& $-$ 	& $-$ 	& 
						\multicolumn{1}{c}{95} &  \\
		\multicolumn{2}{c|}{\textbf{unknown}} & 5 & 5 & $-$ & $-$ & 20 & 30 & $-$ &
						\multicolumn{1}{c}{5} & 
	\end{tabular}
  }
  \caption{Confusion matrix of the recognition rates measured with the CASAS dataset in a simulation performed four times faster than in the real-world. On the top the results for the sequential activity dataset, on the bottom for the interwoven dataset.}
  \label{tab:recognition}
\end{table*}

\subsection{Discussion}
\label{sec:discussion}

All models have been designed via data inspection, with the purpose of identifying the best fitting temporal model thresholds (i.e., $\delta_i$ and $\epsilon_i$), as well as the sensors involved in each specific activity.
Such models describe a \textit{possible} way to recognize the activities performed in the considered dataset, and their design is strongly dependent on both the experimental script and the environmental setup.
Due to the limited number of experiments (a total of $40$ executions), results do not provide an evidence that models accuracy (i.e., recognition rates) are not affected by an over-fitting on the dataset, especially if general ADL such as cooking or cleaning are considered. 

The statement formalization proposed in Section \ref{sec:arianna_computational_framework}, as well as in \autoref{fig:net}, shows the modular capabilities exhibited by Arianna. 
On the one hand, the defined statement's algebra allows for grounding logical rules on an OWL-based ontology via temporal classification.
Its bijective nature allows for considering each model's outcome as a new statement, which is used to update the system's knowledge base through different levels of detail. 
Arianna can be configured to build an on-line reconfigurable network composed of ontologies, which describe human activity recognition models in highly expressive manners, and procedures efficiently triggered by semantic events.
Since events depend on models only, it is possible to have different computations (i.e., network branches) for evaluating the same statement through concurrent sub-models.

As far temporal performance are concerned, \autoref{fig:subPlot} shows that the system can synchronize through the complex flow of a network comprising eight activity models, efficiently validated in parallel through data contextualization.
Via the introduction of a spatial contextualization (i.e., the placing ontology \PlaceOnto{}), we show that the models complexity, and therefore also their reasoning times, can be upper bounded in order to react to events and to recognize activities performed in an interwoven manner. 

\section{Conclusions}

This chapter describes the computational inference engine of Arianna, a smart home system capable of \textit{understanding} whether an assisted person performs a given set of ADL and of \textit{motivating} him/her in performing them through a speech-mediated motivational dialogue.
In so doing, Arianna uses a set of \textit{nearables} to be installed in an apartment, plus a wearable to be worn by the assisted person or fit in garments.
The ideas underlying Arianna are based on new approaches to the management of chronic diseases such as cognitive decline \cite{Bredesen2014,Bredesen2016}, i.e., adopting personalized and multi-therapeutic approaches.
These studies show that adopting a proper lifestyle is an essential step in the management and in some cases the regression of disabling chronic diseases.

The chapter discusses a number of design and implementation choices related to a semantic model able to perform multiple human activity recognition and classification procedures, which has been designed to meet soft real-time requirements in real-world use cases.
In particular, the chapter includes results related to the overall computational capabilities of the architecture. 

Arianna is under active development as a joint effort between University of Genoa and Teseo srl.


\bibliographystyle{spmpsci}      
\bibliography{references}   

\end{document}